\documentclass{article}
\usepackage{src/rl2_arxiv}

\usepackage{amsmath,amsfonts,bm}

\def\eqref#1{equation~\ref{#1}}

\def\1{\bm{1}}

\DeclareMathAlphabet{\mathsfit}{\encodingdefault}{\sfdefault}{m}{sl}
\SetMathAlphabet{\mathsfit}{bold}{\encodingdefault}{\sfdefault}{bx}{n}

\usepackage{graphicx}
\usepackage{booktabs}
\usepackage{url}
\usepackage{amsmath,amssymb,amsthm}
\usepackage{algorithm}
\usepackage{algorithmic}
\usepackage{xcolor}
\usepackage{microtype}
\usepackage{multirow}
\usepackage{hyperref}
\usepackage{placeins}
\usepackage{needspace}
\hypersetup{colorlinks=true,linkcolor=blue!45!black,citecolor=blue!45!black,urlcolor=blue!45!black}
\newcommand{\sg}{\operatorname{sg}}

\title{BeJEPA: A Goal-Conditioned Humanoid\\Behavior World Model via Unsupervised RL}

\author{%
Liyun Yan\textsuperscript{1,2}\quad
Haoyang Li\textsuperscript{1,2}\quad
Yang Zhang\textsuperscript{1}\quad
Jialing Zhu\textsuperscript{1}\quad
Zhanxiang Cao\textsuperscript{1,2}\\
Zihao Zhou\textsuperscript{1,2}\quad
Keqi Zhu\textsuperscript{2,3}\quad
Yizhi Chen\textsuperscript{2,4}\quad
Huang Chuxuan\textsuperscript{2,5}\quad
Shengcheng Fu\textsuperscript{2,4}\\
Zhong Jiangwei\textsuperscript{6}\quad
Qiao Sun\textsuperscript{6}\quad
Chen Zhang\textsuperscript{6}\quad
Yue Gao\textsuperscript{1,2,\textdagger}
}

\newcommand{\paperaffiliations}{%
\textsuperscript{1}RL\textsuperscript{2} Lab, Shanghai Jiao Tong University\quad
\textsuperscript{2}Shanghai Innovation Institute\\
\textsuperscript{3}Zhejiang University\quad
\textsuperscript{4}Tongji University\quad
\textsuperscript{5}Fudan University\\
\textsuperscript{6}Robotics Lab, Lenovo Research, Lenovo Group Limited
}
\newcommand{\paperauthornote}{\textsuperscript{\textdagger}Corresponding author}

\hypersetup{
  pdftitle={BeJEPA: A Goal-Conditioned Humanoid Behavior World Model via Unsupervised RL},
  pdfauthor={Liyun Yan; Haoyang Li; Yang Zhang; Jialing Zhu; Zhanxiang Cao; Zihao Zhou; Keqi Zhu; Yizhi Chen; Huang Chuxuan; Shengcheng Fu; Zhong Jiangwei; Qiao Sun; Chen Zhang; Yue Gao},
  pdfsubject={Goal-conditioned humanoid control through unsupervised reinforcement learning},
  pdfkeywords={Humanoid control, world models, unsupervised reinforcement learning, JEPA}
}

\begin{document}

\maketitle

\thispagestyle{rl2firstpage}

\begin{figure}[!ht]
    \centering
    \includegraphics[width=0.94\linewidth]{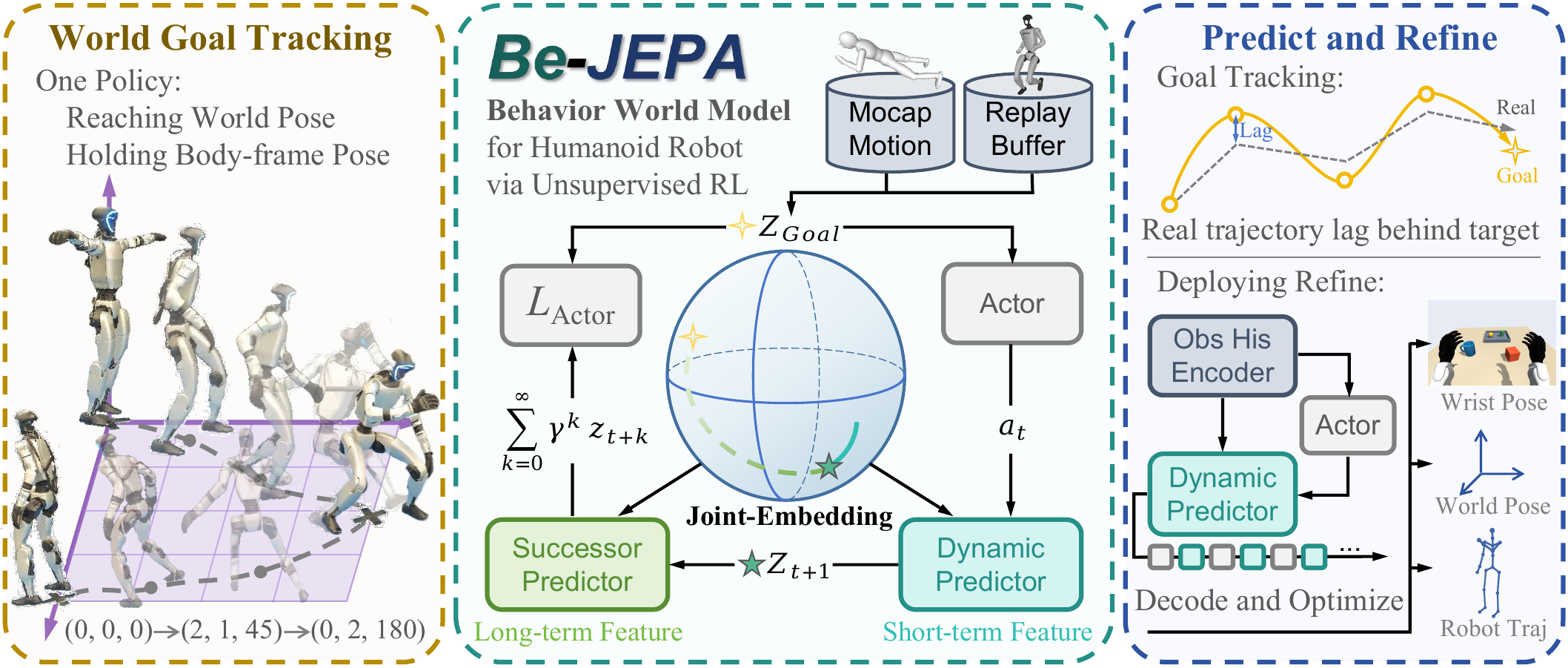}
    \caption{\textbf{BeJEPA: a goal-conditioned Behavior World Model for humanoid control.}
    Left: our goal is to jointly control world position, heading, and whole-body posture.
    Center: Dynamic and Successor learn physical consequences and long-term behavior in a shared latent space through unsupervised RL.
    Right: the Dynamic Predictor refines actions online.}
    \label{fig:teaser}
\end{figure}

\begin{abstract}
Humanoid control requires coordinating whole-body posture with position and heading in the world frame.
Mainstream controllers imitate trajectories in the root frame, but struggle to regain the target world pose after displacement.
We introduce \textbf{BeJEPA}, a goal-conditioned Behavior World Model learned through unsupervised
reinforcement learning from human motion priors and robot interaction.
BeJEPA learns a shared representation of body configuration and world placement
through a joint-embedding predictive architecture (JEPA).
A Dynamic Predictor predicts the next state embedding under an action, while a Successor Predictor
summarizes a goal-conditioned policy's expected discounted future state embeddings in a single vector.
During joint training, Dynamic grounds the initial representation in observed physical transitions,
while Successor increasingly organizes it around long-term behavior.
Their composition guides policy improvement through one physical prediction and one behavioral summary,
without long imagined rollouts.
In 20-second motion-tracking experiments, BeJEPA with online action refinement using Dynamic outperforms
all evaluated baselines in global body-position error, reducing it on AMASS
by 61.7\% relative to BFM-Zero with world-pose inputs and 58.7\% relative to SONIC.
Physical deployment on a Unitree G1 demonstrates joint attainment of posture, world position,
and heading goals, recovery from the ground, and closed-loop return after disturbances and goal changes.
Project page: \url{https://ylyem9x.github.io/BeJEPA-project/}.
\end{abstract}

\clearpage

\section{Introduction}
\label{sec:introduction}
Whole-body control (WBC) is the physical foundation of humanoid applications, enabling manipulation, expressive motion, and the execution of higher-level commands \citep{ben2025homie,wei2026psi0}.
It translates desired robot states into coordinated actuator commands.
For world-coordinate tasks, the desired body posture must be reached together with the requested base position and heading.
Existing motion-centered controllers can reproduce a posture while remaining displaced from its intended location.
Correcting this displacement requires the robot to step, temporarily departing from the requested posture before reaching it at the right place.
Whole-body control must therefore judge a movement by the goal it helps achieve, even when that movement initially worsens posture matching.

Human motion references show coordinated movement, making imitation a practical starting point for learning WBC \citep{peng2018deepmimic,he2025asap,truong2025beyondmimic}.
However, tracking a given trajectory does not explain how to reach a new goal when no trajectory is provided.
This motivates reinforcement learning (RL), through which the robot can discover intermediate movements and evaluate how they help it reach the goal.
Meta Motivo and BFM-Zero use unsupervised RL to learn reusable behaviors \citep{tirinzoni2025metamotivo,li2026bfmzero}.
Given a reward that is linear in the learned state features, these methods can select behaviors to maximize return without specifying intermediate states.
BFM-Zero uses body-relative features, so its learned objective focuses on posture and excludes absolute position and heading.
To control posture and world placement together, the robot must learn how actions change its body and how the resulting behavior brings it closer to the goal.

We introduce \textbf{BeJEPA}, a \emph{goal-conditioned Behavior World Model} for humanoid control, learned through unsupervised RL from human motion priors and robot interaction.
Its shared representation encodes body configuration, world position, and heading.
A goal-independent \emph{Dynamic Predictor} models one-step action consequences, while a goal-conditioned \emph{Successor Predictor} summarizes the policy's expected future state features.
Early in joint training, Dynamic provides physical supervision that helps preserve useful features despite the unskilled policy's noisy long-term targets.
As the policy improves, Successor organizes these features around long-term behavior.
Their composition trains the actor to choose movements by their future goal alignment, including steps that temporarily worsen posture matching, without long imagined rollouts.
Dynamic also supports online action refinement for time-indexed tracking.

Our contributions are threefold:
\begin{itemize}
    \item \textbf{A goal-conditioned Behavior World Model for humanoid control.}
    BeJEPA learns a shared representation of posture and world placement through unsupervised RL, supporting goal reaching and motion tracking.
    \item \textbf{A Dynamic--Successor predictor decomposition.}
    We compose immediate dynamics with a goal-conditioned prediction of future state features, jointly learning the representation and guiding long-term policy improvement without long imagined rollouts.
    \item \textbf{World-anchored control in simulation and on hardware.}
    The learned actor improves global body alignment on AMASS and LAFAN. G1 demonstrates joint posture and world-pose control, ground recovery, and closed-loop return after disturbances and goal changes.
\end{itemize}

\section{Related Work}
\label{sec:related}
\paragraph{Humanoid behavioral foundation models.}
Motion imitation is a dominant route to humanoid behavioral foundation models \citep{peng2018deepmimic,he2025asap,truong2025beyondmimic}.
SONIC and Humanoid-GPT scale tracking accuracy and generalization across motions while retaining a reference-sequence interface \citep{luo2025sonic,qi2026humanoidgpt}.
Isolated or abruptly changed goals thus need an additional mechanism for intermediate motion.
Hierarchical systems supply this motion with a separate planner \citep{tevet2025closd}.
Generated motions must remain within the tracker's capabilities, while replanning frequency trades motion consistency against responsiveness \citep{huang2025diffuse}.
Meta Motivo and BFM-Zero instead infer behaviors directly through unsupervised RL \citep{tirinzoni2025metamotivo,li2026bfmzero}.
However, BFM-Zero's body-relative formulation leaves world position and heading outside its learned goals.
BeJEPA learns world placement through both physical and behavioral prediction, outperforming BFM-Zero retrained with world-pose inputs (Section~\ref{sec:learning-results}).

\paragraph{World models and latent-space control.}
Reconstruction-based models such as PlaNet and Dreamer allocate representation capacity to observation detail, including variation irrelevant to control \citep{hafner2019planet,hafner2020dreamer,hafner2025dreamer}.
Dreamer and the decoder-free TD-MPC family limit rollout horizons and estimate the remaining return with learned values \citep{hafner2020dreamer,hansen2022tdmpc,hansen2024tdmpc2}.
These values depend on training rewards, so their long-term predictions do not automatically transfer to newly specified objectives.
Joint-embedding predictive architectures (JEPAs) learn by predicting state embeddings, allowing abstraction from unpredictable observation detail \citep{assran2023ijepa,assran2025vjepa2}.
Yet planning with latent dynamics, as in DINO-WM and LeWorldModel, composes transitions and risks accumulating prediction error \citep{zhou2024dinowm,maes2026lewm}.
BeJEPA combines feature prediction with a directly learned, goal-conditioned summary of future state features, enabling long-term policy improvement without extended dynamics rollouts or downstream reward supervision.

\paragraph{Successor representations and reward-free reinforcement learning.}
Successor features summarize a policy's expected discounted future state features, separating behavioral prediction from reward readout \citep{barreto2017successor}.
Forward--Backward representations learn this structure to infer policies for new rewards and goals \citep{touati2021fb,touati2023zeroshot}.
TD-JEPA introduces joint-embedding learning of these long-term predictions from offline transitions \citep{bagatella2026tdjepa}.
Acquiring informative transitions online remains a separate problem, motivating dedicated FB exploration strategies \citep{urpi2025epistemically}.
In online humanoid learning, the policy and its long-term prediction targets evolve together.
Early movements may reveal little about how to reach a distant goal, although each executed action already provides an observed physical transition.
BeJEPA learns these two timescales jointly: observed one-step transitions directly supervise the encoder while goal-conditioned continuation organizes the same representation over longer horizons.

\section{BeJEPA: A Behavior World Model}
\label{sec:method}

\begin{figure}[t]
    \centering
    \includegraphics[pagebox=mediabox,trim=8bp 155bp 340bp 27bp,clip,width=0.94\linewidth]{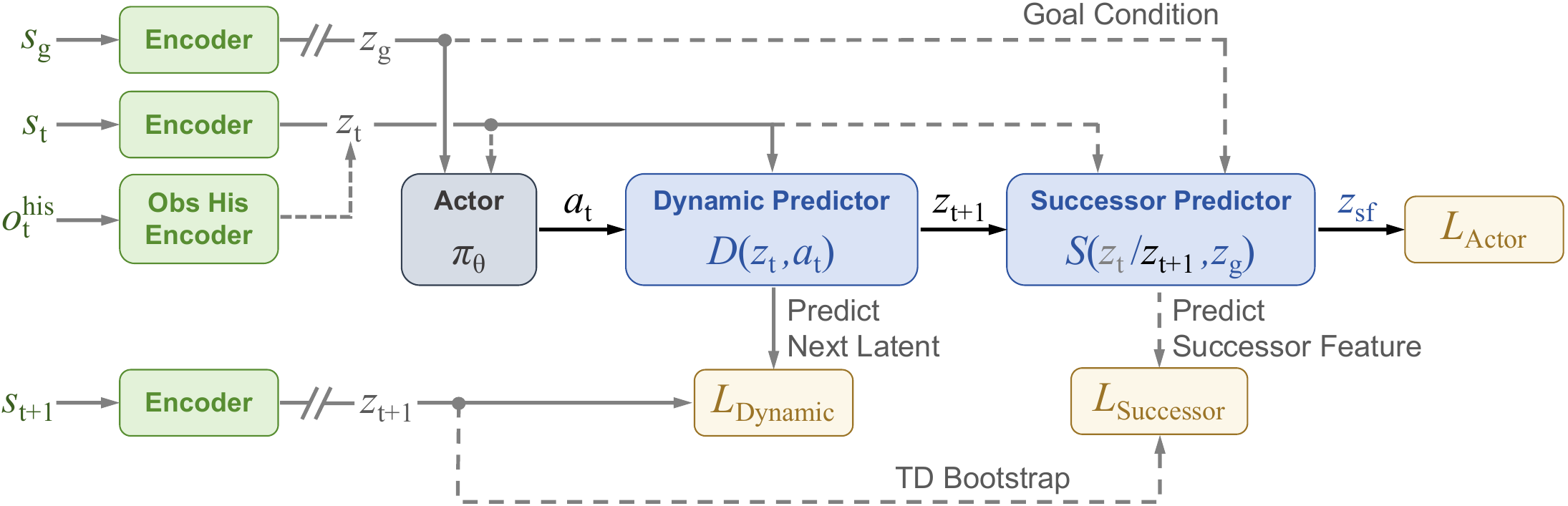}
    \caption{\textbf{Overview of the BeJEPA architecture.}
    Dynamic predicts the observed next latent, while Successor learns discounted continuation under the collection goal.
    Both objectives update the shared encoder.
    The actor learns through their composition, tracing its objective back to the current action.
    Slashed paths denote stop gradient; $z_{\mathrm{sf}}$ denotes a predicted successor feature.}
    \label{fig:architecture}
\end{figure}

BeJEPA learns a controller for goal reaching and motion tracking without task-specific retraining.
It composes Dynamic's one-step physical prediction with Successor's goal-conditioned future-state feature summary in a shared representation of body configuration and world placement to guide policy improvement (Figure~\ref{fig:architecture}).
Dynamic also refines actions online for spatial and temporal alignment.

\subsection{Problem formulation}
\label{sec:representation}

We formulate humanoid control as a partially observable Markov decision process
$\mathcal M=(\mathcal S,\mathcal O,\mathcal A,P,\Omega,p_0,\gamma)$.
Here $s_t\in\mathcal S$ is the full state, $o_t\in\mathcal O$ is the controller's observation, and $a_t\in\mathcal A\subset\mathbb R^{29}$ specifies joint targets for a low-level proportional--derivative controller.
Transitions follow $s_{t+1}\sim P(\cdot\mid s_t,a_t)$, observations follow $\Omega(\cdot\mid s_t)$, initial states follow $p_0$, and $\gamma\in(0,1)$ discounts future outcomes.
The observation contains proprioception, estimated root pose, and recent sensor and action history; the full state additionally exposes simulated body kinematics.
Root pose is represented as $\rho_t=(x_t,y_t,\sin\psi_t,\cos\psi_t)$ relative to a fixed reference anchor shared by current and target states.
A fixed target state $s_g$ specifies goal reaching; a time-indexed reference $s_t^{\mathrm{ref}}$ specifies motion tracking.
We jointly train the world model and a goal-conditioned, off-policy actor--critic from a replay buffer $\mathcal B$ of online robot transitions. 
A retargeted human motion dataset $\mathcal D_{\mathrm{ref}}$ supplies goals and a conditional motion prior \citep{peng2021amp}; this prior and task-independent physical penalties guide executable control without downstream task rewards.

\subsection{Goal-conditioned control with successor features}
\label{sec:successor}

Goal reaching and motion tracking both require matching a target body configuration and its world placement.
Let $E_\phi:\mathcal S\to\mathbb R^d$ encode a state as $z_t=E_\phi(s_t)$ and a target as $z_g=E_\phi(s_g)$.
We normalize both embeddings to radius $\sqrt d$ and define the goal reward by their similarity:
\begin{equation}
    r_g(s)=\langle E_\phi(s),z_g\rangle
    =d-\tfrac12\|E_\phi(s)-z_g\|_2^2.
    \label{eq:goal-reward}
\end{equation}
Thus maximizing reward is exactly minimizing distance to the target in the learned representation.
The goal reward is linear in state features: $z_g$ selects the target, and $E_\phi$ learns the similarity geometry.
A fixed $z_g$ specifies goal reaching; successive reference embeddings specify tracking goals.

Successor features turn instantaneous similarity into a future-behavior objective \citep{barreto2017successor}, providing a basis for zero-shot reward transfer \citep{touati2023zeroshot}.
For policy $\pi_g$ conditioned on $z_g$ and a state region $X\subseteq\mathcal S$, the normalized \emph{successor measure} \citep{touati2021fb} is
\begin{equation}
    M^{\pi_g}(X\mid s)
    =(1-\gamma)\,\mathbb E_{\pi_g}\!\left[
    \sum_{k=0}^{\infty}\gamma^k\mathbf 1\{s_{t+k}\in X\}
    \,\middle|\,s_t=s\right].
    \label{eq:measure}
\end{equation}
This quantity counts discounted visits to $X$, including the current state; the factor $1-\gamma$ makes its total mass one.
Generative temporal-difference models also learn discounted future-state distributions directly, avoiding recursive one-step prediction over the full horizon \citep{janner2020gammamodels}.
Instead of representing a distribution over all future states, we retain its expected embedding:
\begin{equation}
    S^{\pi_g}(s)
    :=\int_{\mathcal S}E_\phi(s')\,M^{\pi_g}(\mathrm ds'\mid s)
    =(1-\gamma)\,\mathbb E_{\pi_g}\!\left[\sum_{k=0}^{\infty}\gamma^k z_{t+k}\,\middle|\,s_t=s\right].
    \label{eq:successor}
\end{equation}
This \emph{successor feature} summarizes the states the policy is expected to occupy.
Because the goal reward is linear in this embedding, its normalized return follows directly:
\begin{equation}
    V_g(s)
    :=\int_{\mathcal S}r_g(s')\,M^{\pi_g}(\mathrm ds'\mid s)
    =\left\langle S^{\pi_g}(s),z_g\right\rangle.
    \label{eq:successor-value}
\end{equation}
The successor predicts the future states visited while pursuing the goal; its inner product with the goal evaluates their alignment.
An action can therefore be useful even if the next posture is less similar to the target, if subsequent behavior brings the robot closer to it.
Contrastive goal-conditioned RL shows a related connection between representation inner products and goal values \citep{eysenbach2022contrastive}.

To learn this long-term prediction from individual transitions, separate the current feature:
\begin{equation}
    S^{\pi_g}(s_t)
    =(1-\gamma)z_t+
    \gamma\,\mathbb E_{a_t\sim\pi_g,\,s_{t+1}\sim P}
    \left[S^{\pi_g}(s_{t+1})\right].
    \label{eq:successor-bellman}
\end{equation}
This vector Bellman equation replaces an entire future trajectory with one observed transition and a prediction of its continuation.
We use this identity to learn a goal-conditioned summary of future state features alongside one-step physical prediction.

\subsection{Learning with Dynamics and Successor Prediction}
\label{sec:dynamic}
\label{sec:actor}

BeJEPA separates learning what an action changes from learning what a goal-conditioned policy subsequently achieves (Figure~\ref{fig:architecture}).
An immediate physical consequence can be learned from every attempted movement, even before the robot can reliably reach the intended goal.
We therefore introduce a goal-independent \emph{Dynamic Predictor} $D_\nu(z_t,a_t)\approx z_{t+1}$ and a goal-conditioned \emph{Successor Predictor} $S_\xi(z_t,z_g)\approx S^{\pi_g}(s_t)$, both operating in the shared representation.
Dynamic reuses physical transitions across goals; Successor shares a state-based continuation across actions with the same latent consequence, reducing the action dependence that must be learned through bootstrapping.

These two predictions connect through the Bellman equation.
Let $Q_g(s_t,a_t)$ be the normalized return obtained by taking $a_t$ now and following $\pi_g$ thereafter.
Substituting Equation~\ref{eq:successor-value} gives
\begin{align}
    Q_g(s_t,a_t)
    &=(1-\gamma)r_g(s_t)+\gamma\,\mathbb E_{s_{t+1}\sim P(\cdot\mid s_t,a_t)}V_g(s_{t+1})\nonumber\\
    &=\left\langle(1-\gamma)z_t+
    \gamma\,\mathbb E_{s_{t+1}\sim P(\cdot\mid s_t,a_t)}S^{\pi_g}(s_{t+1}),\,z_g\right\rangle.
    \label{eq:action-value}
\end{align}
This identity separates the present action from subsequent policy decisions; using a single predicted next latent requires a suitable representation.
Following the JEPA principle of I-JEPA and V-JEPA~2 \citep{assran2023ijepa}, prediction in a learned embedding space encourages retaining predictable structure and abstracting away unpredictable detail.
Here, action conditioning makes a control input's repeatable consequences the prediction target.
The aim is a next-state representation concentrated around a predictable outcome, so that the continuation can be evaluated at $D_\nu(z_t,a_t)$.
This motivates the approximation $\mathbb E[S_\xi(z_{t+1},z_g)\mid s_t,a_t]\approx S_\xi(D_\nu(z_t,a_t),z_g)$ and the composed action value:
\begin{equation}
    \widehat Q_g(z_t,a_t)
    =\big\langle(1-\gamma)z_t+
    \gamma S_\xi(\underbrace{D_\nu(z_t,a_t)}_{\text{action consequence}},z_g),\,z_g\big\rangle.
    \label{eq:composed-value}
\end{equation}
The actor can now evaluate an action through one physical prediction followed by one prediction of the remaining behavior.
The first model is supervised by executed transitions; the second evaluates their longer-term consequences without recursively rolling out that horizon.

Both predictors learn from the robot's own replay transitions.
Let $z_b$ be the goal stored at collection, $d_t\in\{0,1\}$ indicate termination, and $\gamma_t=\gamma(1-d_t)$.
With $\sg$ denoting stop-gradient and $S_{\bar\xi}$ the target network whose parameters $\bar\xi$ track $\xi$ by an exponential moving average, the losses are
\begin{align}
    \mathcal L_D
    &=\mathbb E_{\mathcal B}\!\left[
    \left\|D_\nu(z_t,a_t)-\sg[z_{t+1}]\right\|_2^2\right],\qquad
    \mathcal L_S=\mathbb E_{\mathcal B}\!\left[
    \left\|S_\xi(z_t,z_b)-y_t^S\right\|_2^2\right],
    \label{eq:dynamics}\\
    y_t^S
    &=\sg\!\left[(1-\gamma_t)z_t+\gamma_t S_{\bar\xi}(z_{t+1},z_b)\right].
    \label{eq:successor-target}
\end{align}
The Dynamic loss predicts a detached next-state embedding and updates $E_\phi$ through $z_t$, giving action-predictable information a direct representation-learning signal.
Successor organizes the same coordinates around the future states visited under the stored goal $z_b$.
Its TD target uses the \emph{observed} next state, so training the continuation requires neither a predicted transition nor \mbox{a next-action query to the actor.}
We combine these losses with representation regularization:
\begin{equation}
    \mathcal L_{\mathrm{world}}
    =\lambda_D\mathcal L_D+\mathcal L_S+\mathcal L_{\mathrm{repr}},\qquad
    \mathcal L_{\mathrm{repr}}
    =\lambda_{\mathrm{orth}}\mathcal L_{\mathrm{orth}}
    +\lambda_w\mathbb E_{\mathcal B}\!\left[
    \|H_\zeta^w(z_t)-w(s_t)\|_{W_w}^2\right].
    \label{eq:repr-loss}
\end{equation}
Here $\mathcal L_{\mathrm{orth}}$ discourages redundant features, while a lightweight wrist readout $H_\zeta^w$ preserves fine control detail by reconstructing wrist pose and motion $w(s_t)$; $\|v\|_W^2=v^\top Wv$ denotes a weighted squared error.
The encoder and predictors learn jointly, with $\lambda_D$ decreasing over training to emphasize local physical prediction early and behavioral continuation as the repertoire develops.

An observation-history encoder distills the shared representation into $\widetilde z_t=E_\eta^{\mathrm{obs}}(o_t)$, where $o_t$ includes recent observation history, allowing the actor to operate on control observations and action history $h_t$.
This transfer from privileged training information to deployment observations is also used in teacher--student learning for legged locomotion \citep{miki2022perceptive}.
For goals drawn from $p_g$, which mixes collection goals with replay and motion-reference states, we optimize the composed value in Equation~\ref{eq:composed-value}.
Omitting its action-independent current-state term, the actor loss is
\begin{equation}
    \mathcal L_{\mathrm{actor}}
    =-\gamma\,\mathbb E_{\mathcal B,p_g}\!\left[
    \left\langle S_\xi\!\left(D_\nu\!\left(z_t,\pi_\theta(\widetilde z_t,h_t,z_g)\right),z_g\right),z_g\right\rangle\right]
    +\mathcal R_{\mathrm{act}},
    \label{eq:actor-loss}
\end{equation}
where $\mathcal R_{\mathrm{act}}$ combines the motion prior and physical execution penalties.
Only the actor is updated: gradients pass through the fixed Successor and Dynamic predictors, translating a desirable future state change into a change of the current action.
Algorithm~\ref{alg:training} summarizes training; observation distillation, ensemble aggregation, and regularization details appear in Appendix~\ref{app:training}.

\begin{algorithm}[!htb]
\caption{BeJEPA training}
\label{alg:training}
\small
\begin{algorithmic}
\REQUIRE Motion data $\mathcal D_{\mathrm{ref}}$, replay $\mathcal B$, batch size $n$, goal sampler $p_g$.
\STATE Initialize $E_\phi,D_\nu,S_\xi,H_\zeta^w,E_\eta^{\mathrm{obs}},\pi_\theta$ and target network $S_{\bar\xi}$ with $\bar\xi=\xi$.
\WHILE{interaction budget remains}
    \STATE Collect goal-conditioned transitions into $\mathcal B$.
    \FOR{each replay update}
        \STATE Sample $\{(s_i,o_i,h_i,a_i,s_i^+,z_{b,i},d_i)\}_{i=1}^n\sim\mathcal B$ and encode:
        \STATE $z_i=E_\phi(s_i)$; $z_i^+=E_\phi(s_i^+)$; $\widetilde z_i=E_\eta^{\mathrm{obs}}(o_i)$; $\gamma_i=\gamma(1-d_i)$.
        \STATE $\mathcal L_D=\frac1n\sum_i\|D_\nu(z_i,a_i)-\sg[z_i^+]\|_2^2$.
        \STATE $\mathcal L_S=\frac1n\sum_i\big\|S_\xi(z_i,z_{b,i})-\sg[(1-\gamma_i)z_i+\gamma_i S_{\bar\xi}(z_i^+,z_{b,i})]\big\|_2^2$.
        \STATE $\mathcal L_{\mathrm{world}}=\lambda_D\mathcal L_D+\mathcal L_S+\mathcal L_{\mathrm{repr}}$, with $\mathcal L_{\mathrm{repr}}$ from Equation~\ref{eq:repr-loss}.
        \STATE $\mathcal L_{\mathrm{obs}}=\frac1n\sum_i\|\widetilde z_i-\sg[z_i]\|_2^2$.
        \STATE $\{z_{g,i}\}_{i=1}^n\sim p_g$; $\widehat a_i=\pi_\theta(\widetilde z_i,h_i,z_{g,i})$.
        \STATE $\mathcal L_{\mathrm{actor}}=-\frac\gamma n\sum_i\langle S_\xi(D_\nu(z_i,\widehat a_i),z_{g,i}),z_{g,i}\rangle+\mathcal R_{\mathrm{act}}$.
        \STATE Minimize $\mathcal L_{\mathrm{world}}$ over $(E_\phi,D_\nu,S_\xi,H_\zeta^w)$ and $\mathcal L_{\mathrm{obs}}$ over $E_\eta^{\mathrm{obs}}$.
        \STATE Update only $\pi_\theta$ to minimize $\mathcal L_{\mathrm{actor}}$, keeping encoders and predictors fixed.
        \STATE Update target parameters $\bar\xi\leftarrow(1-\tau)\bar\xi+\tau\xi$.
    \ENDFOR
\ENDWHILE
\end{algorithmic}
\end{algorithm}

\subsection{Aligning Behavior in Space and Time}
\label{sec:mpc}

Motion tracking requires the right body configuration \emph{at the right place and time}.
Successive goal embeddings guide the actor, but its objective evaluates discounted future similarity to each goal, leaving room for spatial offset and temporal lag.
Finite-dimensional task features can also blur spatial precision, as analyzed in FB-AWARE \citep{cetin2024finer}.
We use Dynamic to compare a short predicted trajectory with time-indexed references and refine the actor's actions online, using gradient-based latent MPC similar to Dream-MPC \citep{spieler2026dreammpc}.

The actor and planner start from the same observation-history embedding $\widehat z_0=\widetilde z_t=E_\eta^{\mathrm{obs}}(o_t)$, without privileged current-state inputs.
Reference embeddings $z_{t+1:t+H}^{\mathrm{ref}}$ come from the commanded motion.
For bounded action residuals $\delta_k$, the rollout is
\begin{equation}
    \bar a_k=\pi_\theta(\widehat z_k,h_k,z_{t+k+1}^{\mathrm{ref}}),\quad
    a_k=\operatorname{clip}_{\mathcal A}(\bar a_k+\delta_k),\quad
    \widehat z_{k+1}=\mathcal D_{t,k}(\widehat z_k,a_k),
    \label{eq:mpc-rollout}
\end{equation}
where $h_k$ advances the action history and $\mathcal D_{t,k}$ uses the frozen Dynamic ensemble, optionally corrected by recent observation--prediction discrepancies (Appendix~\ref{app:mpc}).
Only the first $R\leq H$ residuals are optimized; the remaining steps follow the actor.

A root-pose readout trained on detached latents and wrist readout expose geometric errors alongside latent similarity.
Writing $e_{\mathrm{geom},k}(z)$ for root-position, wrapped-heading, and wrist-pose reference errors at step $k$, define
$c_k(z)=\lambda_z\|z-z_{t+k+1}^{\mathrm{ref}}\|_2^2/d+\|e_{\mathrm{geom},k}(z)\|_W^2$.
All execution priorities use:
\begin{equation}
    \mathcal J_t(\delta)
    =\sum_{k=0}^{H-1}\gamma^k\!\left[
    c_k(\widehat z_{k+1})+
    \lambda_a\|a_k-\bar a_k\|_2^2+
    \lambda_\Delta\|a_k-a_{k-1}\|_2^2\right],
    \qquad \|\delta_k\|_\infty\leq\epsilon,
    \label{eq:mpc-cost}
\end{equation}
where $a_{-1}$ is the last executed action.
Cost weights, planning horizons, and adjustable joints specify desired precision.
A bounded gradient search compares corrections with the actor rollout, executes the first action of the lowest-cost sequence, and replans from the next observation.
Reference-dependent activation can retain the actor outside the refinement regime.
All network weights remain fixed: the physical model learned for behavior acquisition also supports online spatial and temporal alignment.

\Needspace{12\baselineskip}
\section{Experiments}
\label{sec:experiments}
Our experiments examine whether BeJEPA can coordinate body posture with world placement, and how physical and behavioral prediction support this ability.
We ask four questions:
\begin{list}{}{\setlength{\leftmargin}{2.4em}\setlength{\labelwidth}{2.1em}\setlength{\labelsep}{.3em}}
\item[\textbf{Q1.}] How well does BeJEPA perform on standard tracking metrics?
\item[\textbf{Q2.}] Does the Dynamic--Successor split improve world-goal control?
\item[\textbf{Q3.}] How are position, motion, and physical state organized in the latent space?
\item[\textbf{Q4.}] How does BeJEPA perform on the physical robot?
\end{list}

\subsection{Q1: How well does BeJEPA track whole-body motion?}
\label{sec:tracking-results}
We pretrain with $4{,}096$ parallel Isaac Sim environments, following massively parallel legged RL \citep{rudin2022walk}.
We evaluate the 29-actuator G1 policy in MuJoCo at 50\,Hz on selected non-interactive AMASS segments \citep{mahmood2019amass} and LAFAN segments \citep{harvey2020lafan}, each lasting 20\,s to assess long-distance tracking.
Unlike prior protocols that exclude trajectories with large spatial deviations and mark them as failures, we retain these trajectories to fairly evaluate the error in world coordinates.
We measure pelvis-centered and global MPJPE, root translation and heading errors, and local wrist position and orientation errors.
We compare with SONIC, Humanoid-GPT, and BFM-Zero retrained with world-pose inputs.

\begin{table}[!htbp]
\centering\small
\setlength{\tabcolsep}{2.5pt}
\caption{\textbf{Whole-body tracking.} MPJPE is in mm. Root err.\ is $e_{xy}+0.01e_{\mathrm{yaw}}$ (m, degrees); wrist err.\ is $e_{\mathrm{pos},L}+0.5e_{\mathrm{rot},L}$ (cm, degrees). Bold and underline indicate the best and second-best results.}
\label{tab:tracking}
\begin{tabular}{@{}lrrrr@{\hspace{10pt}}rrrr@{}}
\toprule
 & \multicolumn{4}{c}{AMASS} & \multicolumn{4}{c}{LAFAN} \\
\cmidrule(lr){2-5}\cmidrule(lr){6-9}
Method & MPJPE$_G\!\downarrow$ & MPJPE$_L\!\downarrow$ & \shortstack{Root\\err.$\downarrow$} & \shortstack{Wrist\\err.$\downarrow$} & MPJPE$_G\!\downarrow$ & MPJPE$_L\!\downarrow$ & \shortstack{Root\\err.$\downarrow$} & \shortstack{Wrist\\err.$\downarrow$} \\
\midrule
BFM-Zero (world) & 573.4 & 97.1 & 0.801 & 32.4 & 797.1 & 89.3 & 1.020 & 25.8 \\
SONIC & 531.5 & \textbf{49.7} & 0.571 & \textbf{15.3} & 887.5 & \underline{77.8} & 0.971 & \underline{23.6} \\
Humanoid-GPT & 832.4 & 115.7 & 1.107 & 27.5 & 1433.1 & 127.9 & 1.720 & 29.7 \\
\midrule
BeJEPA & \underline{278.0} & 82.3 & \underline{0.377} & 25.9 & \underline{584.5} & 81.9 & \underline{0.716} & 24.3 \\
BeJEPA-MPC (Balanced) & \textbf{219.5} & \underline{63.9} & \textbf{0.255} & \underline{16.3} & \textbf{543.8} & \textbf{75.3} & \textbf{0.639} & \textbf{20.1} \\
\bottomrule
\end{tabular}
\end{table}

\paragraph{World placement is part of the behavior goal.}
Local posture accuracy can remain high despite drift from the reference path, making global alignment essential to world-directed control.
BeJEPA has the lowest global MPJPE and root error among the learned policy baselines on both datasets (Table~\ref{tab:tracking}).
The shared goal specifies desired body pose and its world placement.
Compared with SONIC, BeJEPA improves world placement despite higher MPJPE$_L$.
Correcting world position can require an extra step that sacrifices local pose matching, making MPJPE$_G$ better suited to world-coordinate evaluation, especially as targets derive from human trajectories infeasible for robots.

\begin{figure}[!htbp]
\centering
\includegraphics[width=0.90\linewidth]{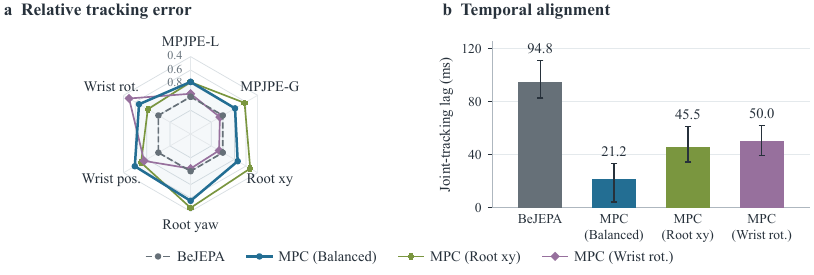}
\caption{\textbf{Tracking trade-offs across execution priorities and temporal alignment on AMASS.}
(a) Six errors normalized by BeJEPA (1.0); outward is better on a common linear scale.
(b) Mean joint-tracking lag over $\pm500$\,ms shifts (20\,ms steps).}
\label{fig:mpc-tradeoffs}
\end{figure}

\paragraph{Physical prediction aligns execution with the reference clock.}
BeJEPA's successor objective values continued proximity to a goal; tracking additionally requires arriving at each reference state at its assigned time.
BeJEPA-MPC (Balanced) reduces error across all six metrics, with mean relative reductions of $31.2$\% on AMASS and $13.4$\% on LAFAN using the same pretrained weights.
We examine temporal alignment by shifting each trajectory's reference to minimize joint-angle error.
BeJEPA-MPC (Balanced) reduces the mean signed lag from $94.8$ to $21.2$\,ms on AMASS (Figure~\ref{fig:mpc-tradeoffs}b) and from $97.2$ to $51.0$\,ms on LAFAN.
This connects Dynamic's two roles: transition supervision learns control-relevant consequences, and online prediction uses them to align behavior with time-indexed targets.
This addresses the temporal precision required beyond future goal similarity (Section~\ref{sec:mpc}).

\paragraph{Execution priorities can be specified at inference.}
At deployment, Dynamic supports task-specific precision through the MPC cost and rollout settings.
BeJEPA-MPC (Root xy) and BeJEPA-MPC (Wrist rot.) emphasize root translation and wrist orientation, respectively, improving the corresponding metric with trade-offs elsewhere (Figure~\ref{fig:mpc-tradeoffs}a).
The same frozen model can thus adapt execution to a task's priorities without retraining (Appendix~\ref{app:mpc}).

\subsection{Q2: Does the Dynamic--Successor split improve world control?}
\label{sec:learning-results}
We compare BeJEPA with world-input BFM-Zero, a merged predictor of comparable capacity, split variants with either predictor's encoder gradient stopped, and Dynamic-only MPC.
The gradient ablations retain both predictors and the actor objective, isolating each predictor's contribution to the representation.
Under fixed goals, minimum position and heading errors measure closest approach, while mean errors measure alignment throughout execution.

\begin{table}[!htbp]
\centering\small
\setlength{\tabcolsep}{6pt}
\caption{\textbf{Dynamic--Successor ablation on goal inference.} Lower errors are better.}
\label{tab:ablation}
\begin{tabular}{lrrrr}
\toprule
 & \multicolumn{2}{c}{Root $xy$ (m)} & \multicolumn{2}{c}{Root yaw ($^\circ$)} \\
\cmidrule(lr){2-3}\cmidrule(lr){4-5}
Model & Min. & Mean & Min. & Mean \\
\midrule
BFM-Zero (world) & 0.450 & 0.984 & 8.28 & 46.26 \\
Merged predictor & 0.666 & 1.348 & 6.41 & 35.81 \\
Split, no Dynamic encoder gradient & 0.645 & 1.231 & 4.76 & 34.10 \\
Split, no Successor encoder gradient & 0.221 & 0.885 & 3.19 & 25.41 \\
Dynamic-only MPC & 0.675 & 1.004 & 11.30 & 82.23 \\
\midrule
BeJEPA & \textbf{0.159} & \textbf{0.642} & \textbf{2.47} & \textbf{24.02} \\
\bottomrule
\end{tabular}
\end{table}

\paragraph{Both predictors contribute to world control.}
The merged successor predictor has large world-goal errors (Table~\ref{tab:ablation}), consistent with the difficulty of learning from noisy long-term targets.
Splitting improves alignment even with Dynamic's encoder gradient stopped, supporting the structural benefit in unsupervised RL, although errors remain large.
Training the encoder through Dynamic while stopping Successor's gradient improves alignment further, but joint training performs best.
Dynamic grounds the features; Successor organizes them for long-term control.
MPC following LeWorldModel~\citep{maes2026lewm} without the actor also struggles with distant goals: motion exceeds Dynamic's reliable local horizon (Appendix~\ref{app:predictive-validation}), motivating Successor's long-term summary without extended rollouts.

\subsection{Q3: What structure does BeJEPA give the latent space?}
\label{sec:geometry-results}
\begin{figure}[!htb]
\centering
\includegraphics[width=0.92\linewidth]{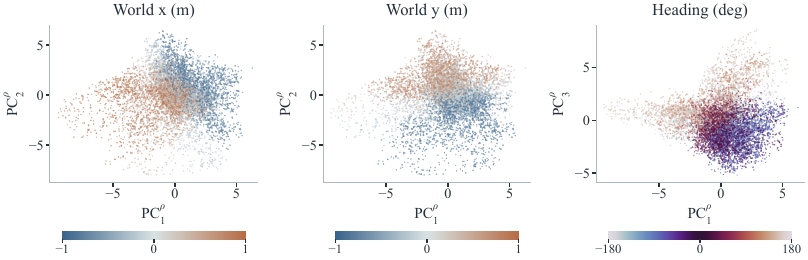}
\caption{\textbf{World geometry within a shared body-state representation.}
Eighty body states are crossed with random world poses: $x,y\in[-1,1]$\,m and yaw in $[-\pi,\pi]$.
Within-body PCA (PC$^\rho$) subtracts body means before projection; colors show $x$, $y$, and heading.}
\label{fig:world-geometry}
\end{figure}

\paragraph{Body content and world placement jointly define a goal.}
We randomly sample body states from the buffer and cross them with randomized $x,y,$ and yaw values.
To expose how world placement changes a goal, we subtract each body's mean embedding across world poses and fit a shared principal component analysis (PCA) to these changes.
The resulting coordinates vary systematically with translation and heading (Figure~\ref{fig:world-geometry}).
World $x$ varies mainly along PC$_1^\rho$ and world $y$ along PC$_2^\rho$, while PC$_3^\rho$ partially separates heading, with substantial overlap.
The representation retains body content while distinguishing its possible world placements, supplying the latent goal reward's geometry.

\begin{figure}[!htb]
\centering
\includegraphics[width=0.94\linewidth]{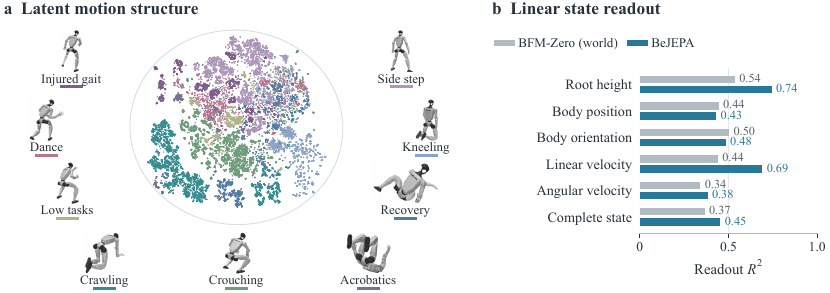}
\caption{\textbf{Motion organization and linear state readouts in the latent space.}
(a) t-SNE with root $x=y=\psi=0$; colors match the illustrated motion families.
(b) Linear state readouts at the original world poses.}
\label{fig:motion-readouts}
\label{fig:physical-readouts}
\end{figure}

\paragraph{A common world pose reveals the whole-body repertoire.}
We examine trajectories in the motion prior, placing their body states at a common root position and heading.
The t-SNE map organizes crawling, kneeling, upright gaits, and expressive movement into characteristic posture neighborhoods (Figure~\ref{fig:motion-readouts}a).
Similar body poses tend to occupy nearby regions in this projection, suggesting that the latent space captures meaningful posture similarity across motions.

\paragraph{Physical quantities are more linearly accessible.}
Linear readouts measure accessible physical information (Figure~\ref{fig:physical-readouts}b).
Relative to world-input BFM-Zero, BeJEPA improves linear readout of body velocity and root height, while position and orientation remain comparable.
Although lower scores do not rule out nonlinear recovery, these gains, together with the posture neighborhoods and encoder-gradient ablation, suggest that BeJEPA learns a more structured latent representation, with physical information more readily accessible to linear readouts.

\subsection{Q4: How does BeJEPA perform on the physical robot?}
\label{sec:deployment-results}
We deploy the pretrained BeJEPA policy on a physical Unitree G1.
The onboard LiDAR, coupled with FAST-LIO~\citep{xu2021fastlio,xu2022fastlio2}, provides the position and heading estimates used by the world-anchored observation interface.
Together with proprioception, these estimates allow the policy to relate the robot's current body configuration to a goal placed in the world.

\begin{figure}[!htbp]
\centering
\includegraphics[width=\linewidth]{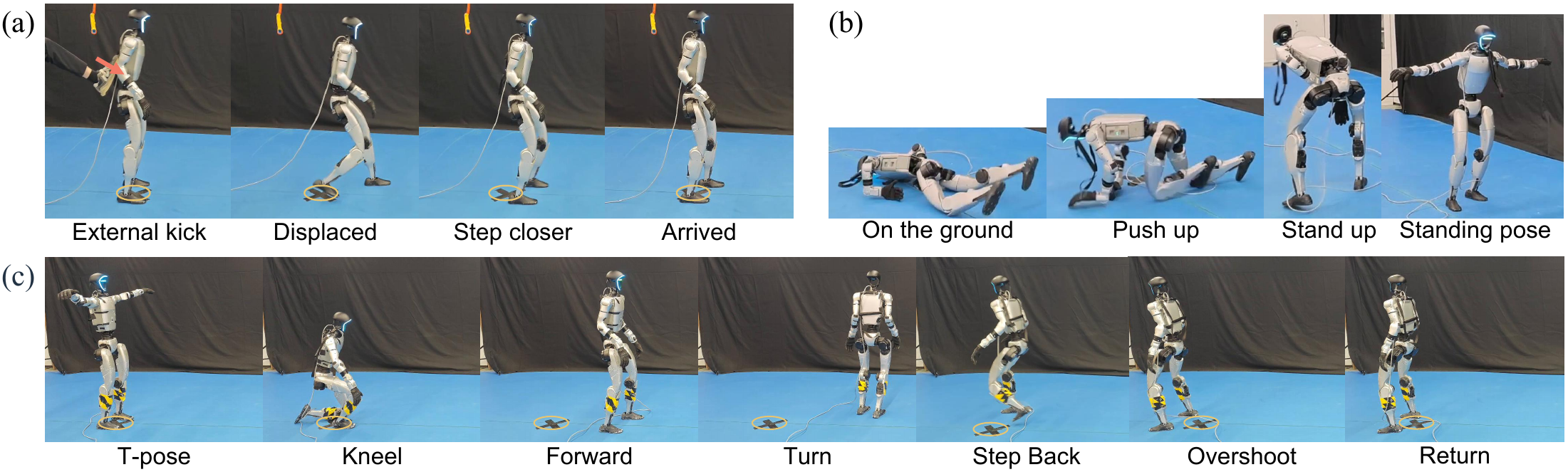}
\caption{\textbf{World-anchored control on the physical G1.}
(a) Recovery after a kick. (b) Standing up from the ground. (c) Sequential goal switching.
Gold rings identify the floor reference marks in (a,c).}
\label{fig:real-deployment}
\end{figure}

\paragraph{Recovering balance and posture.}
A world-anchored goal remains meaningful when an external disturbance moves the robot away from it.
Following a kick, the G1 first steps toward its original standing location, then makes a smaller second step to complete the return (Figure~\ref{fig:real-deployment}a). Starting from the ground presents a larger change in body configuration: the robot pushes up, rises, reorients, and reaches the commanded standing pose (Figure~\ref{fig:real-deployment}b). The two sequences demonstrate complementary aspects of recovery: correcting displacement and restoring an upright body configuration.

\FloatBarrier

\paragraph{Changing behavior preserves the spatial reference.}
Sequential goals take the robot from a T-pose through kneeling, advancing, and turning before returning to its starting location (Figure~\ref{fig:real-deployment}c).
Returning to the same floor mark despite these posture and heading changes demonstrates that the learned behaviors retain a persistent spatial reference.

\medskip

\subsection*{AI use statement}
Generative AI tools assisted with manuscript drafting, polishing, code analysis, and preparation of analysis and plotting scripts.
We have reviewed all AI-assisted work, including code, text, data, and claims. We take responsibility for the final content of this work.

\bibliography{src/references}
\bibliographystyle{plainnat}

\clearpage
\appendix
\section*{Appendix}
\begingroup

\section{Successor Measures and Temporal-Difference Latent Prediction}
\label{app:derivation}

This section develops the successor-feature interpretation used in Section~\ref{sec:successor} and connects it to TD-JEPA~\citep{bagatella2026tdjepa}.
Throughout, $\pi_g$ denotes a fixed goal-conditioned policy, $E_\phi$ a fixed feature map when deriving prediction identities, and $s$ the Markov state including the controller history.
We use the normalized, current-state convention of Equation~\ref{eq:measure}.

\subsection{A distribution over future states}
Let $K$ be independent of the trajectory with $\Pr(K=k)=(1-\gamma)\gamma^k$, $k\geq0$.
The successor measure has the probabilistic interpretation
\begin{equation}
    M^{\pi_g}(X\mid s)=\Pr(s_{t+K}\in X\mid s_t=s,\pi_g).
    \label{eq:app-geometric-time}
\end{equation}
It is therefore a probability measure over states visited at a geometrically distributed future time.
Writing $P_g(\mathrm ds^+\mid s)=\int_{\mathcal A}P(\mathrm ds^+\mid s,a)\pi_g(\mathrm da\mid s)$, separating the event $K=0$ gives
\begin{equation}
    M^{\pi_g}(\mathrm ds'\mid s)
    =(1-\gamma)\delta_s(\mathrm ds')+
    \gamma\int_{\mathcal S}M^{\pi_g}(\mathrm ds'\mid s^+)P_g(\mathrm ds^+\mid s).
    \label{eq:app-measure-bellman}
\end{equation}
Integrating $E_\phi(s')$ yields the vector Bellman equation
\begin{equation}
    S^{\pi_g}(s)=(1-\gamma)E_\phi(s)+\gamma\mathbb E_{s^+\sim P_g(\cdot\mid s)}S^{\pi_g}(s^+).
    \label{eq:app-bellman}
\end{equation}
For a finite state space, let $\Phi$ contain the state features as rows and $\mathsf S_g$ their successor features. Then
\begin{equation}
    \mathsf S_g=(1-\gamma)(I-\gamma P_g)^{-1}\Phi
    =(1-\gamma)\sum_{k\geq0}\gamma^kP_g^k\Phi.
    \label{eq:app-resolvent}
\end{equation}
The predictor thus compresses repeated application of the policy's transition operator into a single feature vector.

\subsection{From future-feature regression to TD-JEPA}
Consider predicting $Y=E_\phi(s_{t+K})$ from the current state and goal.
Its conditional mean is exactly $S^{\pi_g}(s)$.
For any prediction $f(s,z_g)$, expanding the square around this mean gives
\begin{align}
    \mathbb E\!\left[\|f(s,z_g)-Y\|_2^2\mid s,g\right]
    &=\|f(s,z_g)-S^{\pi_g}(s)\|_2^2\nonumber\\
    &\quad+\mathbb E\!\left[\|Y-S^{\pi_g}(s)\|_2^2\mid s,g\right].
    \label{eq:app-mc-regression}
\end{align}
The cross term vanishes because $\mathbb E[Y-S^{\pi_g}(s)\mid s,g]=0$.
The second term is independent of $f$: regression onto sampled future embeddings learns their discounted barycenter.
This is the latent-prediction interpretation of successor features in TD-JEPA.

Temporal-difference learning replaces the sampled future by the one-step decomposition in Equation~\ref{eq:app-bellman}.
Define $\mathcal T_gF=(1-\gamma)E_\phi+\gamma P_gF$.
For bounded vector-valued functions,
\begin{equation}
    \|\mathcal T_gF-\mathcal T_gG\|_\infty
    \leq\gamma\|F-G\|_\infty.
    \label{eq:app-contraction}
\end{equation}
Hence $S^{\pi_g}$ is its unique fixed point, and repeated Bellman updates recover the same expected feature as future-state regression.
In BeJEPA, the sampled regression target is
\begin{equation}
    y_t^S=\sg\!\left[(1-\gamma_t)E_\phi(s_t)
    +\gamma_tS_{\bar\xi}(E_\phi(s_{t+1}),z_b)\right],
    \qquad \gamma_t=\gamma(1-d_t),
    \label{eq:app-td-target}
\end{equation}
where $z_b$ is the collection goal and $d_t$ indicates termination.
Both features are computed by the online encoder, and the target predictor follows
$\bar\xi\leftarrow(1-\tau)\bar\xi+\tau\xi$.
The stored goal associates the observed transition with its behavioral continuation; goal relabeling for policy improvement is described in Appendix~\ref{app:goals}.
At termination, the target reduces to the current feature.

The action-conditioned predictor in TD-JEPA uses future features starting at the next state.
In normalized notation, write this quantity as
$F_g^+(s,a)=(1-\gamma)\mathbb E[\sum_{k\geq0}\gamma^kE_\phi(s_{t+k+1})\mid s_t=s,a_t=a,\pi_g]$.
The two conventions are related by
\begin{equation}
    F_g^+(s,a)=\mathbb E_{s^+\mid s,a}S^{\pi_g}(s^+),\qquad
    S^{\pi_g}(s)=(1-\gamma)E_\phi(s)+\gamma\mathbb E_{a\sim\pi_g}F_g^+(s,a).
    \label{eq:app-conventions}
\end{equation}
BeJEPA represents the first expression through a physical transition followed by a goal-conditioned continuation, $S_\xi(D_\nu(z,a),z_g)$.

\subsection{Why the prediction supports control}
For a linear feature reward $r_w(s)=E_\phi(s)^\top w$, integration against the successor measure gives
\begin{equation}
    (1-\gamma)\mathbb E_{\pi_g}\!\left[\sum_{k\geq0}\gamma^k r_w(s_{t+k})\mid s_t=s\right]
    =S^{\pi_g}(s)^\top w.
    \label{eq:app-linear-return}
\end{equation}
Here $g$ determines the behavior and $w$ determines the reward readout.
BeJEPA chooses $w=z_g=E_\phi(s_g)$.
Because $\|E_\phi(s)\|_2^2=\|z_g\|_2^2=d$, the resulting goal value also satisfies
\begin{equation}
    V_g(s)=d-\frac{1-\gamma}{2}\mathbb E_{\pi_g}\!\left[
    \sum_{k\geq0}\gamma^k\|E_\phi(s_{t+k})-z_g\|_2^2\mid s_t=s\right].
    \label{eq:app-distance-return}
\end{equation}
Policy improvement therefore minimizes distance to the goal across the discounted future.
To connect this objective to the present action, the composition in Equation~\ref{eq:composed-value} gives
\begin{equation}
    \nabla_\theta\widehat Q_g
    =\gamma\left(\frac{\partial\pi_\theta}{\partial\theta}\right)^{\!\top}
    \left(\frac{\partial D_\nu}{\partial a}\right)^{\!\top}
    \left(\frac{\partial S_\xi}{\partial z^+}\right)^{\!\top}z_g,
    \qquad z^+=D_\nu(z_t,a_t).
    \label{eq:chain}
\end{equation}
The current-state reward has zero action derivative.
Successor supplies a direction of improvement in the next-state representation, and Dynamic translates that direction into an action update.
The encoders and predictors are held fixed while this gradient updates the actor.

\subsection{Why separating Dynamic and Successor improves learning}
\label{app:factorization}
We show how separating action consequences from behavioral continuation improves replay reuse while preserving the goal-conditioned control objective.

Following JEPA's emphasis on predictable features \citep{assran2023ijepa,assran2025vjepa2}, assume that $z=E_\phi(s)$ removes unpredictable nuisance variation while retaining all information needed for goal rewards and policy decisions.
The resulting latent is Markov, and its next state is determined by the current latent and action: $z^+=D^\star(z,a)$.
Fix the encoder and a goal-conditioned policy $\pi_g$ for the following argument.

\begin{proof}
Let $S_g(z)$ denote the policy's successor feature in this representation.
By Equation~\ref{eq:app-conventions}, the action-conditioned future feature satisfies
\begin{equation}
    F_g^+(z,a)=\mathbb E[S_g(z^+)\mid z,a]
    =S_g(D^\star(z,a)).
    \label{eq:app-exact-factorization}
\end{equation}
Substituting this identity into Equation~\ref{eq:action-value} recovers the composed value in Equation~\ref{eq:composed-value} with exact predictors.
The decomposition therefore preserves action values and their maximizing actions.

The same identity exposes the sharing that makes learning more efficient.
A merged predictor learns continuation as a function of $(z,a,z_g)$.
The split assigns immediate action dependence to Dynamic, whose observed next-latent targets can be learned from transitions collected under any goal.
Each improvement in this physical prediction is consequently reused across goals.
Successor learns continuation as a function of $(z,z_g)$: all predecessor state--action pairs reaching the same latent share its future-feature prediction.
An update to that continuation therefore benefits every action leading to it, reducing duplicated long-term prediction across actions.
Moreover, Equation~\ref{eq:successor-target} learns this continuation directly from observed state transitions, without querying a next action in the bootstrap.
Thus action effects receive direct one-step supervision, while TD learning estimates a shared state-based continuation; the long-term predictor no longer needs to separately learn the initiating action's effect on future value.
\end{proof}

Appendix~\ref{app:learning} supports this efficiency gain empirically: even with Dynamic's encoder gradient stopped, the split model achieves lower early global-tracking and fixed-goal heading errors than the merged predictor.
Full BeJEPA improves further when Dynamic also shapes the encoder, linking the predictable-representation premise to the observed learning advantage.

\section{Model-Predictive Action Refinement}
\label{app:mpc}

\subsection{A shared prediction and refinement objective}
All MPC settings use the same frozen actor, History Encoder, Dynamic ensemble, and geometric readouts.
The History Encoder initializes the current latent from deployable observations; the state encoder supplies reference-goal embeddings.
The actor provides coordinated nominal behavior, while Dynamic predicts how small action changes alter its execution.
We average the two predictors' costs and propagate their mean latent, reducing sensitivity to an individual predictor's errors.
The stage cost in Equation~\ref{eq:mpc-cost} is
\begin{equation}
    c_k(z)=\frac{\lambda_z}{d}\|z-z_{t+k+1}^{\mathrm{ref}}\|_2^2
    +\lambda_{xy}\|e_{xy,k}\|_2^2+\lambda_\psi e_{\psi,k}^2
    +\frac{\lambda_{wp}}{6}\|e_{wp,k}\|_2^2
    +\frac{\lambda_{wr}}{12}\|e_{wr,k}\|_2^2.
    \label{eq:app-mpc-stage}
\end{equation}
Root and wrist positions are in meters, heading errors are wrapped to $[-\pi,\pi]$, and each wrist orientation uses a six-dimensional rotation representation.
The latent term preserves whole-body agreement, while geometric readouts make task-specific precision explicit.
Scoring every predicted step with $\gamma=0.98$ constrains the intermediate motion as well as its destination; residual and smoothness penalties retain continuity with the actor.

\subsection{Execution priorities within the same controller}
Table~\ref{tab:mpc-settings} gives the settings reported in Section~\ref{sec:tracking-results}.
Balanced refinement combines latent, root, and wrist terms over a horizon long enough to anticipate coordinated motion.
Root-$xy$ refinement increases the translation weight and prediction horizon, allowing earlier actions to account for their later effect on world placement.
Wrist-orientation refinement instead concentrates the cost on the hands and restricts corrections to the arm and wrist joints over a short horizon.
This directs the action adjustment toward local precision while retaining the actor's lower-body behavior.
These priorities change how the same physical predictions are evaluated, without learning a new policy for each task.

\begin{table}[!htbp]
\centering\small
\caption{MPC settings for the reported execution priorities. $H$: prediction horizon; $R$: residual horizon; $N$: gradient updates. Bounds $\epsilon$ and step sizes $\eta$ are in normalized action coordinates. All settings use per-step costs with $\gamma=0.98$.}
\label{tab:mpc-settings}
\begin{tabular}{lrrrrrrr}
\toprule
Setting & $H$ & $R$ & $N$ & $\epsilon$ & $\eta$ & $\lambda_a$ & $\lambda_\Delta$ \\
\midrule
Balanced & 16 & 8 & 8 & 0.20 & 0.080 & 0.10 & 0.030 \\
Root $xy$ & 24 & 8 & 8 & 0.24 & 0.080 & 0.03 & 0.010 \\
Wrist rotation & 3 & 3 & 6 & 0.30 & 0.150 & 0.01 & 0.005 \\
\bottomrule
\end{tabular}

\medskip
\begin{tabular}{lrrrrrl}
\toprule
Setting & $\lambda_z$ & $\lambda_{xy}$ & $\lambda_\psi$ & $\lambda_{wp}$ & $\lambda_{wr}$ & Refined joints \\
\midrule
Balanced & 20 & 80 & 12 & 300 & 30 & All \\
Root $xy$ & 20 & 120 & 12 & 0 & 0 & All \\
Wrist rotation & 0.2 & 0.5 & 0.1 & 600 & 100 & Arms and wrists \\
\bottomrule
\end{tabular}
\end{table}

Balanced and root-$xy$ refinement apply to reference segments with mean planar speed at most $0.8$\,m/s, retaining the actor on faster segments; wrist refinement applies throughout.
This confines whole-body intervention to slower reference motions while allowing local hand corrections across the motion set.
The activation rule uses the supplied reference, so a task's motion requirements determine when refinement is applied.

\paragraph{Filtering prediction bias for world placement.}
Root-$xy$ refinement uses the supplied reference root poses directly and anchors predicted displacement to the currently observed root pose.
This stabilizes world coordinates, a low-frequency component that is prone to overshoot.
To reduce persistent transition bias, we also filter the discrepancy between observed and predicted latents:
\begin{align}
    b_t&=(1-\alpha)b_{t-1}+\alpha\big[\widetilde z_t-\overline D_\nu(\widetilde z_{t-1},a_{t-1})\big],\nonumber\\
    \widehat z_{k+1}^{(i)}&=\Pi\!\left(D_\nu^{(i)}(\widehat z_k,a_k)+\beta^k b_t\right),
    \label{eq:app-mpc-correction}
\end{align}
where $\overline D_\nu$ is the ensemble mean, $\Pi$ projects onto the latent sphere, $\alpha=0.5$, and $\beta=0.9$.
Exponential averaging suppresses rapid fluctuations in the discrepancy while retaining a persistent offset; through a locally linear root readout, it acts approximately as filtering the position innovation.
The horizon decay limits how far the current correction is extrapolated.
This causal adjustment uses past actions and observations to improve the rollout estimate while preserving the shared tracking objective and action search.
Balanced and wrist refinement use the uncorrected predictions and geometric targets decoded from reference embeddings.

\section{Training and Network Architecture}
\label{app:training}

\subsection{Motion data and goal relabeling}
\label{app:data}
Our motion prior contains $2{,}000$ G1 motion clips: $862$ ten-second LAFAN clips~\citep{harvey2020lafan} and $1{,}138$ motions from the SONIC/Bones collection~\citep{luo2025sonic}.
Together they provide approximately $4.78$ hours of motion at 30\,Hz, covering upright locomotion, crouching, crawling, kneeling, recovery, and expressive whole-body movement.
Retargeted body states provide reference goals and expert state--goal pairs for the conditional prior; robot interaction supplies the state--action transitions used to learn Dynamic and Successor.

\label{app:goals}
Half of the rollout environments follow moving reference goals in 250-step windows, with randomized initial position and heading.
The remaining environments collect fixed-goal behavior, including translations of up to 3\,m and heading changes of up to $\pi$.
Ground initializations occur with probability $0.3$, providing experience for posture recovery.

Replay retains the collection goal $z_b$ for Successor TD learning.
For actor and CPR/AUX critic updates, we replace this goal with probability $0.8$.
The replacement is an encoded replay state with probability $0.4$ and an encoded motion-reference state with probability $0.6$.
Thus the marginal training-goal distribution is
\begin{equation}
    p_g=0.20\,p_{\mathrm{behavior}}+0.32\,p_{\mathrm{replay}}+0.48\,p_{\mathrm{reference}}.
    \label{eq:app-goal-mixture}
\end{equation}
Replay candidates are independently reassigned by a batch permutation.
This exposes policy improvement to attained states and reference behaviors while retaining goals encountered during collection.

\subsection{Shared representation and core parameters}
The shared latent has dimension $d=256$ and radius $\sqrt d$.
Table~\ref{tab:architecture} lists the network architecture and parameter counts.
The state encoder, History Encoder, Dynamic, Successor, and actor use residual multilayer perceptrons; Dynamic and Successor each have two ensemble members.
The History Encoder learns the shared coordinates by distillation:
\begin{equation}
    \mathcal L_{\mathrm{obs}}=\mathbb E_{\mathcal B}\!\left[
    \|E_\eta^{\mathrm{obs}}(o_t)-\sg[E_\phi(s_t)]\|_2^2\right].
    \label{eq:distill}
\end{equation}
The actor combines this embedding, the goal, and recent action history to produce 29 joint commands.
The training encoder receives proprioception, body kinematics, and root pose; the History Encoder receives proprioception, estimated root pose, and recent observation history, supplying the current latent for both the actor and Dynamic at deployment.

\begin{table}[!t]
\centering\small
\caption{Network architecture and size. Depth denotes the main-network hidden-layer count. The CPR/AUX critics use a two-branch design similar to BFM-Zero~\citep{li2026bfmzero}; each input branch adds four residual blocks, explaining their larger parameter count than Successor. Parameter counts include all ensemble members and exclude target-network copies and normalization buffers. M denotes one million parameters.}
\label{tab:architecture}
\begin{tabular}{lrrrr}
\toprule
Network & Width & Depth & Members & Parameters (M) \\
\midrule
State / goal encoder $E_\phi$ & 1,024 & 6 & 1 & 7.12 \\
History Encoder $E_\eta^{\mathrm{obs}}$ & 512 & 4 & 1 & 1.44 \\
Dynamic $D_\nu$ & 2,048 & 3 & 2 & 27.44 \\
Successor $S_\xi$ & 2,048 & 6 & 2 & 53.57 \\
Actor $\pi_\theta$ & 2,048 & 6 & 1 & 31.64 \\
Wrist readout $H_\zeta^w$ & 256 & 4 & 1 & 0.27 \\
Root-pose readout $H_\zeta^\rho$ & 128 & 2 & 1 & 0.05 \\
CPR critic $C$ & 2,048 & 6 & 2 & 134.98 \\
AUX critic $A$ & 2,048 & 6 & 2 & 134.98 \\
Motion-prior discriminator & 1,024 & 3 & 1 & 2.91 \\
\midrule
\multicolumn{4}{l}{Core model (encoders, predictors, actor, and readouts)} & 121.54 \\
\multicolumn{4}{l}{All online networks, including training guidance} & 394.41 \\
\multicolumn{4}{l}{Actor and History Encoder only} & 33.08 \\
\bottomrule
\end{tabular}
\end{table}

The feature regularizer in Equation~\ref{eq:repr-loss} is
\begin{equation}
    \mathcal L_{\mathrm{orth}}=-\frac1B\sum_i\|z_i\|_2^2
    +\frac{1}{2B(B-1)}\sum_{i\neq j}(z_i^\top z_j)^2,
    \label{eq:app-orth}
\end{equation}
which encourages distinct feature directions under sphere normalization.
Since $\|z_i\|_2^2=d$ after projection onto the sphere, the first term is the constant $-d$ and is omitted in computation.
A wrist head reconstructs root-relative wrist position, orientation, and velocity with coefficient $0.05$, preserving fine motion information in the shared representation.
A root-pose head is fitted to detached latents with coefficient $0.05$ and supplies position and heading readouts for MPC.

\begin{table}[!t]
\centering\small
\caption{Core pretraining parameters.}
\label{tab:optimization}
\begin{tabular}{ll@{\qquad}ll}
\toprule
Parameter & Value & Parameter & Value \\
\midrule
Parallel environments & $4{,}096$ & Replay capacity & $5.12\times10^6$ \\
Batch size & $4{,}096$ & Updates / collection step & $16$ \\
Discount $\gamma$ & $0.98$ & Target EMA $\tau$ & $0.005$ \\
Encoder learning rate & $10^{-5}$ & Predictor learning rate & $3\times10^{-4}$ \\
Actor learning rate & $3\times10^{-4}$ & Observation encoder LR & $3\times10^{-4}$ \\
Critic learning rate & $10^{-4}$ & Discriminator LR & $10^{-5}$ \\
Orthogonality weight & $10^{-4}$ & Wrist reconstruction weight & $0.05$ \\
CPR / AUX coefficients & $0.005$ each & Actor exploration std. & $0.1$ \\
Optimizer & Adam & Gradient norm limit & $10$ \\
\bottomrule
\end{tabular}
\end{table}

Each collection step adds $4{,}096$ transitions, followed by 16 replay updates.
Dynamic uses one-step prediction throughout training.
Its loss weight is $0.1$ for the first 30 million transitions and follows a cosine decay to $0.001$ at 100 million transitions.
The encoder learning rate similarly decays from $10^{-5}$ to $10^{-7}$ between 50 and 150 million transitions.
These schedules emphasize learning physical consequences early, then give behavioral continuation increasing influence as the representation stabilizes.

\subsection{CPR and AUX guidance}
Conditional policy regularization (CPR) follows the motion-prior approach of BFM-Zero~\citep{li2026bfmzero}.
A discriminator learns to distinguish reference state--goal pairs from collected state--goal pairs using logistic classification with a gradient penalty.
Its reward trains a goal-conditioned critic $C$ that favors behavior compatible with the motion prior.
An auxiliary critic $A$ learns a physical-execution reward combining torque, action smoothness, joint-limit, contact, foot-orientation, and slippage penalties.
Both critics use the Bellman target
\begin{equation}
    y_t^j=r_t^j+\gamma_t\overline Q^{\,j}(s_{t+1},z_g,a_{t+1}),
    \qquad j\in\{\mathrm{CPR},\mathrm{AUX}\},\quad a_{t+1}\sim\pi_\theta(\cdot\mid o_{t+1},z_g),
    \label{eq:app-regularizer-targets}
\end{equation}
where $Q^{\mathrm{CPR}}=C$ and $Q^{\mathrm{AUX}}=A$; the physical reward is scale-normalized.

For scalar ensemble predictions $q_1,\ldots,q_m$, define
\begin{equation}
    \operatorname{Pess}_\kappa(q)=\frac1m\sum_iq_i-
    \frac{\kappa}{m(m-1)}\sum_{i\ne j}|q_i-q_j|.
    \label{eq:app-pessimism}
\end{equation}
The two Dynamic and two Successor members give four composed values
$q_{ij}=\langle(1-\gamma)z_t+\gamma S_j(D_i(z_t,a_t),z_g),z_g\rangle$.
Using $\kappa=0.25$ for this ensemble and for the two-member critics, the complete actor objective is
\begin{equation}
    \mathcal L_\pi=-\frac1d\mathbb E\!\left[q^-+
    w\left(0.005C^-+0.005A^-\right)\right],
    \qquad w=\sg\!\left[\mathbb E|q^-|\right],
    \label{eq:app-full-actor}
\end{equation}
where $^-$ denotes pessimistic aggregation.
The detached scale $w$ keeps the prior guidance proportional to the learned goal value.
CPR promotes coherent movement and AUX shapes physical execution, complementing the goal objective supplied by Dynamic and Successor.

\section{Learning Dynamics}
\label{app:learning}

\subsection{Convergence of posture and world alignment}
Q2 (Section~\ref{sec:learning-results}) compares the final learned behaviors; Figure~\ref{fig:app-convergence} follows their acquisition throughout training.
We compare BeJEPA with the merged predictor and two split models that stop either Dynamic's or Successor's gradient to the encoder, evaluating their actors directly under matched tracking and fixed-goal protocols.

Local posture tracking improves quickly across models, while global tracking and fixed-goal errors expose differences in learning world alignment.
The merged predictor must learn immediate action effects and goal-conditioned continuation together, without a separately supervised physical transition model.
The split model with Dynamic's encoder gradient stopped benefits from separating these predictions, but its representation receives no direct physical-prediction gradient; world alignment remains slower and less accurate than in full BeJEPA.

When only Dynamic among the two predictors updates the encoder, convergence is generally better than with the merged predictor or stopped Dynamic gradient.
This ablation retains a trained Successor for policy improvement, but removes its influence on the representation.
Its remaining fixed-goal position and heading errors indicate that physical predictability alone is insufficient for accurate longer-term goal reaching.
Full BeJEPA combines physical supervision with goal-conditioned continuation learning in the shared encoder, yielding faster world alignment and lower residual goal errors.

\begin{figure}[!htbp]
\centering
\includegraphics[width=\linewidth]{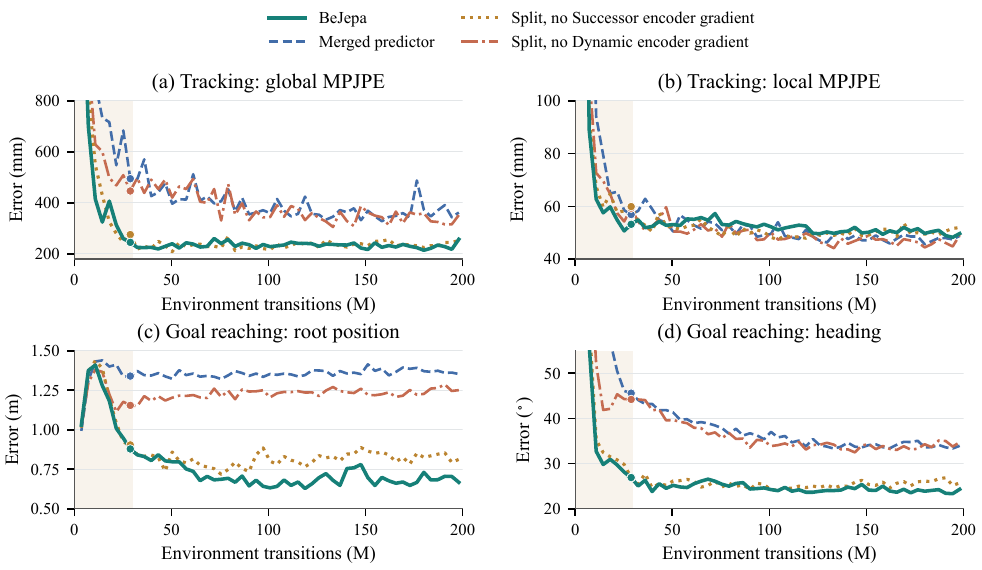}
\caption{\textbf{Tracking and fixed-goal errors over training.}
(a,b) Global and local tracking MPJPE.
(c,d) Mean position and heading errors from the separate fixed-goal evaluation.
The two split ablations stop the indicated predictor's gradient to the encoder while retaining both predictors; shading and markers highlight early learning.}
\label{fig:app-convergence}
\end{figure}

\subsection{How Dynamic and Successor shape the encoder}
Let $g_D=\nabla_\phi\mathcal L_D$ and $g_S=\nabla_\phi\mathcal L_S$ denote the encoder gradients of the unweighted prediction losses.
We examine their directional alignment and their contributions after applying the training weights:
\begin{equation}
    c_{DS}=\frac{\langle g_D,g_S\rangle}{\|g_D\|_2\|g_S\|_2},
    \qquad G_D=\lambda_D\|g_D\|_2,\qquad G_S=\|g_S\|_2.
    \label{eq:app-gradient-diagnostics}
\end{equation}
These quantities are computed on the shared encoder, where physical prediction and behavioral continuation influence the same coordinates.

\paragraph{Early learning: physical prediction establishes the latent coordinates.}
During the first 5 million transitions, the median weighted gradient ratio $G_D/G_S$ is $2.63$; during 5--10 million it is $1.02$, as Successor catches up (Figure~\ref{fig:app-gradients}a--c).
Dynamic therefore supplies the dominant initial prediction signal, with the two contributions becoming comparable within the first 10 million transitions.
Its prediction error drops rapidly over this interval.
Observed action consequences can thus shape the encoder before reliable goal-conditioned behavior has formed, providing the physical coordinates in which longer-term continuation is subsequently learned.

\paragraph{Intermediate learning: behavioral continuation builds on physical structure.}
After roughly 15 million transitions, Successor becomes the larger prediction gradient: the median $G_D/G_S$ is $0.39$ during 15--30 million and $0.20$ during 30--100 million.
Meanwhile, mean gradient cosine similarity rises from $0.119$ during 10--15 million to $0.160$ during 15--30 million and $0.226$ during 30--100 million (Figure~\ref{fig:app-gradients}a,b).
The change in magnitude therefore accompanies greater directional agreement: Successor increasingly organizes the representation around long-term behavior, while Dynamic continues to constrain it through predictable physical change.
For a small step along the Dynamic gradient, holding prediction targets fixed,
\begin{equation}
    \mathcal L_S(\phi-\alpha g_D)-\mathcal L_S(\phi)
    =-\alpha\langle g_S,g_D\rangle+O(\alpha^2).
    \label{eq:app-gradient-interaction}
\end{equation}
Positive alignment identifies a shared local descent direction for the two objectives.
Its increase during this stage supports the interpretation that predictable action consequences provide useful representational structure for learning behavioral continuation.

\paragraph{Late learning: the objectives settle with unequal influence.}
After 100 million transitions, when $\lambda_D$ has decayed to $0.001$ and the control curves approach a plateau, the median $G_D/G_S$ falls to $0.0031$ and mean cosine similarity falls to $0.069$ (Figure~\ref{fig:app-gradients}a--c).
All logged interval-averaged cosines remain positive, and both prediction errors settle at low values.
Together, these observations are consistent with physical prediction establishing useful coordinates early, the two objectives jointly shaping long-term representations in the intermediate stage, and Successor dominating later refinement as behavior stabilizes.

\begin{figure}[!htbp]
\centering
\includegraphics[width=\linewidth]{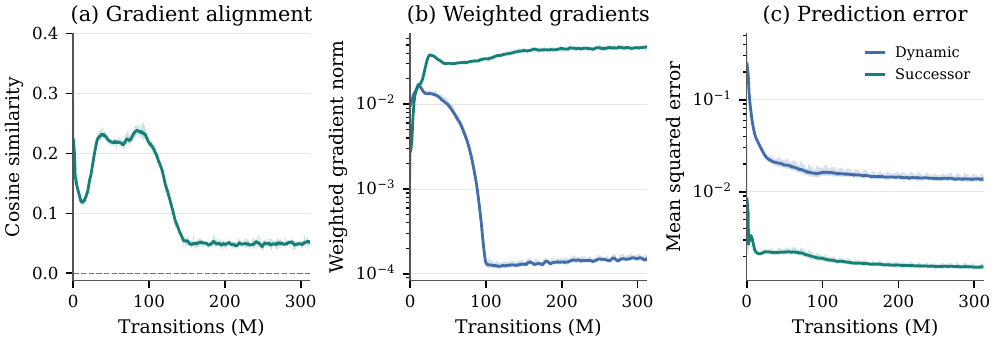}
\caption{\textbf{Encoder-gradient alignment, weighted norms, and prediction losses.}
(a) Gradient cosine similarity on the shared encoder.
(b) Gradient norms after applying loss coefficients.
(c) Unweighted prediction losses averaged over latent coordinates and ensemble members.
Faint lines show averages over each logging interval; solid lines show a nine-record moving average.
Panels (b,c) use logarithmic vertical axes and share the legend in (c).}
\label{fig:app-gradients}
\end{figure}
\section{Additional Experiments}
\label{app:additional-experiments}

\subsection{Acquiring position, heading, and a simple posture together}
\label{app:joint-goals}
Can the policy jointly reach a world position and a requested body posture?
We command standing, arms-out, and shallow-squat goals at displaced positions and headings, and track root-position, yaw, and joint-angle errors throughout the approach and subsequent holding period.
Figure~\ref{fig:app-simple-goals} resolves the initial adjustment densely and follows the same executions after arrival.

\begin{figure}[!htbp]
\centering
\includegraphics[width=\linewidth]{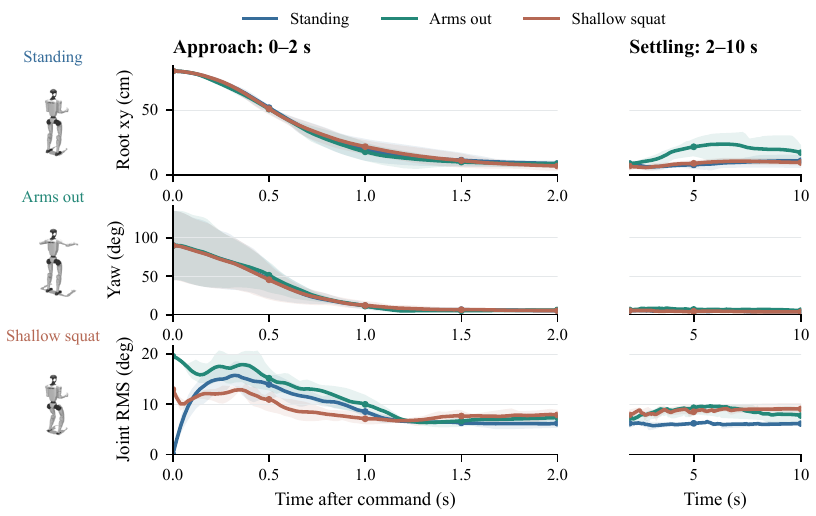}
\caption{\textbf{Root-position, heading, and joint-angle errors during fixed-goal execution.}
Lines show mean errors; shading shows the interquartile range.
The early approach is expanded horizontally, with markers at $0$, $0.5$, $1$, $1.5$, $2$, $5$, and $10$\,s.
The illustrated postures receive the same position and heading commands.}
\label{fig:app-simple-goals}
\end{figure}

The initial joint-error response differs across postures: it rises from zero for standing, but falls for arms out and shallow squat.
This is consistent with a shared need to adopt a moving configuration before settling into the final pose.
In particular, remaining exactly in the standing target would prevent the leg motion needed to walk.
As the robot approaches the destination, position and heading errors decrease together, while joint errors leave the initial locomotion transient and enter a lower, bounded range.
The resulting behavior is a gradual transition from traveling toward the goal to adjusting the requested posture near it.
It does not require a manually specified switch between walking and posing.

\begin{table}[!htbp]
\centering\small
\caption{\textbf{Simultaneous goal acquisition (\%).}
Cumulative fraction satisfying root-position error $\leq0.2$\,m, yaw error $\leq15^\circ$, and joint RMS $\leq10^\circ$ together for $0.5$\,s.
This measures acquisition; the later curves in Figure~\ref{fig:app-simple-goals} show holding behavior.}
\label{tab:app-simple-goals}
\begin{tabular}{lrrrrrrr}
\toprule
Time (s) & 0 & 0.5 & 1 & 1.5 & 2 & 5 & 10 \\
\midrule
Standing & 0.0 & 0.0 & 0.0 & 12.5 & 87.5 & 100.0 & 100.0 \\
Arms out & 0.0 & 0.0 & 0.0 & 0.0 & 87.5 & 100.0 & 100.0 \\
Shallow squat & 0.0 & 0.0 & 0.0 & 25.0 & 75.0 & 100.0 & 100.0 \\
\bottomrule
\end{tabular}
\end{table}

Table~\ref{tab:app-simple-goals} confirms that the three requirements are attained simultaneously, rather than in separate executions.
After acquisition, the errors remain bounded near the goal, although residual motion persists, especially for arms out.
Together, the approach and holding curves support coordinated world alignment and posture control; they do not imply exact, motionless convergence.

\subsection{How far can Dynamic predict without a new observation?}
\label{app:predictive-validation}
How long does Dynamic remain useful when no new state observation is available?
Starting from an encoded hardware state, we recursively propagate the executed actions,
\begin{equation}
    \widehat z_0=E_\phi(s_t),\qquad
    \widehat z_{k+1}=D_\nu(\widehat z_k,a_{t+k}),
    \label{eq:app-open-loop}
\end{equation}
and compare the predicted features and their root-pose readouts with the measured future states.
No intermediate observation corrects the rollout.

\begin{figure}[!htbp]
\centering
\includegraphics[width=\linewidth]{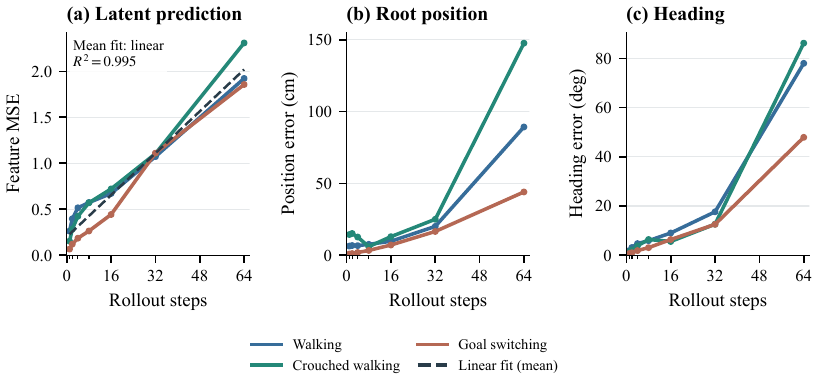}
\caption{\textbf{Feature and root-pose prediction errors versus Dynamic rollout horizon.}
Markers evaluate $1,2,4,8,16,32,$ and $64$ steps, with $20$\,ms per step.
(a) Feature MSE; the dashed line fits the mean across executions as a linear function of elapsed time.
(b,c) Root-position and heading errors, including readout error.
Each colored curve averages over starting points within an execution.}
\label{fig:app-real-dynamic}
\end{figure}

The average feature error grows approximately linearly over the measured range; a quadratic fit offers little additional explanatory power.
The physical readouts remain local over roughly the first eight steps ($160$\,ms), then show more pronounced error growth at longer horizons: the forecasts eventually drift substantially in both position and heading.
Thus the one-step predictor should not be interpreted as a uniformly accurate long-range simulator.

These results support treating Dynamic as a local model of action consequences.
In BeJEPA, the subsequent behavior is represented by Successor, which learns discounted future features directly from observed transitions.

\subsection{Separating destination from arrival heading}
\label{app:goal-paths}
Can the robot reach destinations in different directions while independently controlling its arrival heading?
We place stationary posture goals on a circle around the initial base position.
In one condition, target headings point radially outward; in the other, every destination shares the same world heading.
The resulting paths and heading vectors distinguish control of world position from simply walking forward along the body's facing direction.

\begin{figure}[!htbp]
\centering
\includegraphics[width=0.9\linewidth]{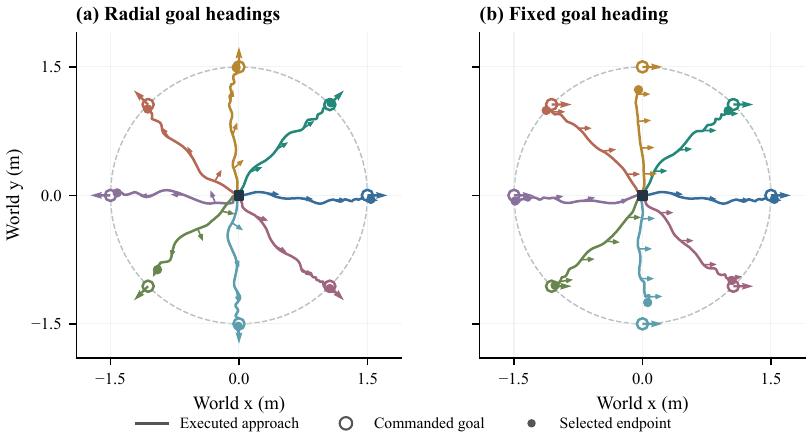}
\caption{\textbf{Planar root trajectories and base headings for circularly arranged goals.}
Targets lie on a $1.5$\,m circle. Open markers and arrows denote target position and yaw; filled markers denote the displayed endpoints.
Short, equal-length arrows show base heading along the path, not velocity.
We show the arms-out posture with the lowest mean combined alignment error across conditions, tracing each path to its best simultaneous position, heading, and joint alignment within the evaluation window.}
\label{fig:app-goal-paths}
\end{figure}

With radial targets, heading vectors turn toward the requested outward orientation as the paths approach the circle.
With a fixed target heading, the robot reaches destinations to its sides and behind it while arriving with a common orientation.
Travel direction and arrival heading therefore need not coincide.
This supports the role of the world-position and yaw components in goal conditioning: the policy can adjust translation and orientation jointly instead of tying each destination to a single forward-facing approach.
The figure illustrates the approach to the selected endpoint; sustained holding is assessed separately in Figure~\ref{fig:app-simple-goals}.
\clearpage
\endgroup

\end{document}